%% file: gpl_iclr2022_conference.tex
\documentclass{article} 
\usepackage{gpl_iclr2022_conference,times}

\input{math_commands.tex}

\usepackage{hyperref}
\usepackage{url}
\usepackage{cite}
\usepackage{amsmath,amssymb,amsfonts}
\usepackage{algorithmic}
\usepackage{graphicx}
\usepackage{textcomp}
\usepackage{float}
\usepackage{natbib}
\usepackage{subcaption}
\usepackage{booktabs}
\usepackage{blindtext}

\title{Let’s Handle It: Generalizable Manipulation of Articulated Objects}

\author{
Zhutian Yang$^*$ \& Aidan Curtis$^*$ \\
Department of Electrical Engineering and Computer Science\\
Massachusetts Institute of Technology\\
Cambridge, MA 02139, USA \\
\texttt{\{curtisa, ztyang\}@mit.edu}
}

\iclrfinalcopy 
\begin{document}

\maketitle


\begin{abstract}
In this project we present a framework for building generalizable manipulation controller policies that map from raw input point clouds and segmentation masks to joint velocities. We took a traditional robotics approach, using point cloud processing, end-effector trajectory calculation, inverse kinematics, closed-loop position controllers, and behavior trees. We demonstrate our framework on four manipulation skills on common household objects that comprise the SAPIEN ManiSkill Manipulation challenge. The source code for this project can be found at \url{https://github.com/aidan-curtis/az-manipulator}
\end{abstract}

\section{Introduction}

A fundamental capability of household robots is manipulation of furniture such as cabinets, doors, chairs, and heavy objects. These skills differ from rigid-body manipulation which has been a typical focus of manipulation research~\citep{graspnet, BodnarRSS20, kpam}. A recent benchmark and corresponding challenge called the SAPIEN ManiSkill Challenge~\citep{sapien_challenge, mu2021maniskill}\footnote{https://sapien.ucsd.edu/challenges/maniskill2021/} aims to evaluate the quality and generalizability of such manipulation policies. The challenge involves manipulating a diverse set of articulated 3D objects that are commonly found in households inside the SAPIEN~\citep{Xiang_2020_SAPIEN} simulation engine. Specifically, the four tasks are:
\begin{enumerate}
    \item \textit{Open Cabinet Drawer} exemplifies motions constrained by a revolute joint. A designated drawer needs to be fully open by a single-arm robot.
    \item \textit{Open Cabinet Door} exemplifies motions constrained by a prismatic joint. A designated door needs to be fully open ($90^{\circ}$ change of joint position) by a single-arm robot.
    \item \textit{Push Chair} exemplifies motions constrained by an object's contact with the plane. A chair needs to arrive at the target location in a vertical pose by a dual-arm robot.
    \item \textit{Move Bucket} exemplifies motions with constraints. A bucket, which contains a ball that moves freely inside, needs to arrive at the target location by a dual-arm robot.
\end{enumerate}
In this project, we developed a framework that uses the traditional robotics approach, including point cloud processing, end-effector trajectory calculation, inverse kinematics, closed-loop position controllers, and behavior trees. After describing our method in detail, we will evaluate our method, analyze the failure cases, and finally discuss the pros and cons of our method and the challenge as a whole.
\begin{figure}[hbt]
    \centering
    \begin{subfigure}[b]{0.22\textwidth}
        \includegraphics[width=\textwidth]{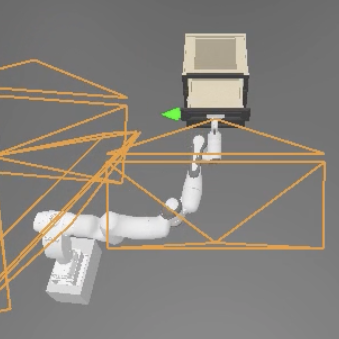}
        \caption{Open drawer}
    \end{subfigure} \hfill 
    \begin{subfigure}[b]{0.22\textwidth}
        \includegraphics[width=\textwidth]{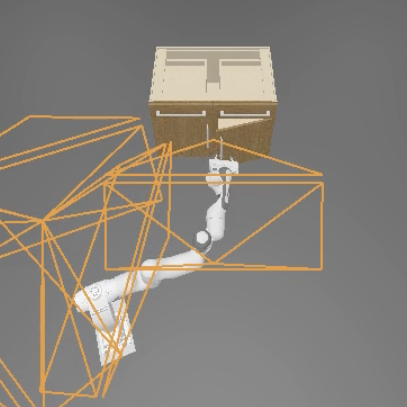}
        \caption{Open door}
    \end{subfigure}  \hfill 
    \begin{subfigure}[b]{0.22\textwidth}
        \includegraphics[width=\textwidth]{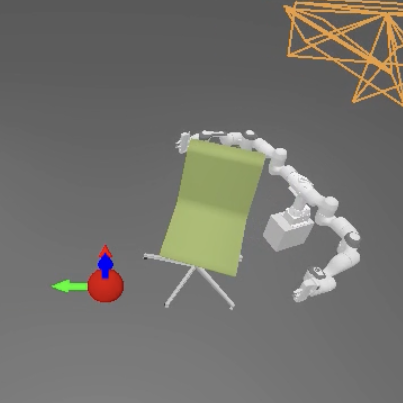}
        \caption{Push chair}
    \end{subfigure} \hfill 
    \begin{subfigure}[b]{0.22\textwidth}
        \includegraphics[width=\textwidth]{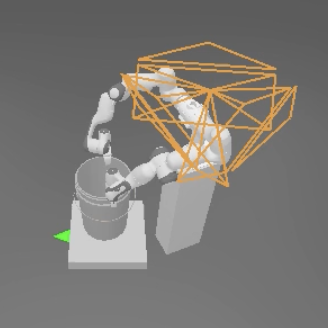}
        \caption{Move bucket}
    \end{subfigure}
    \caption{The four tasks in ManiSkill challenge}
\end{figure}
\begin{figure}[h]
    \centering
     \begin{subfigure}[b]{0.44\textwidth}
        \includegraphics[width=\textwidth]{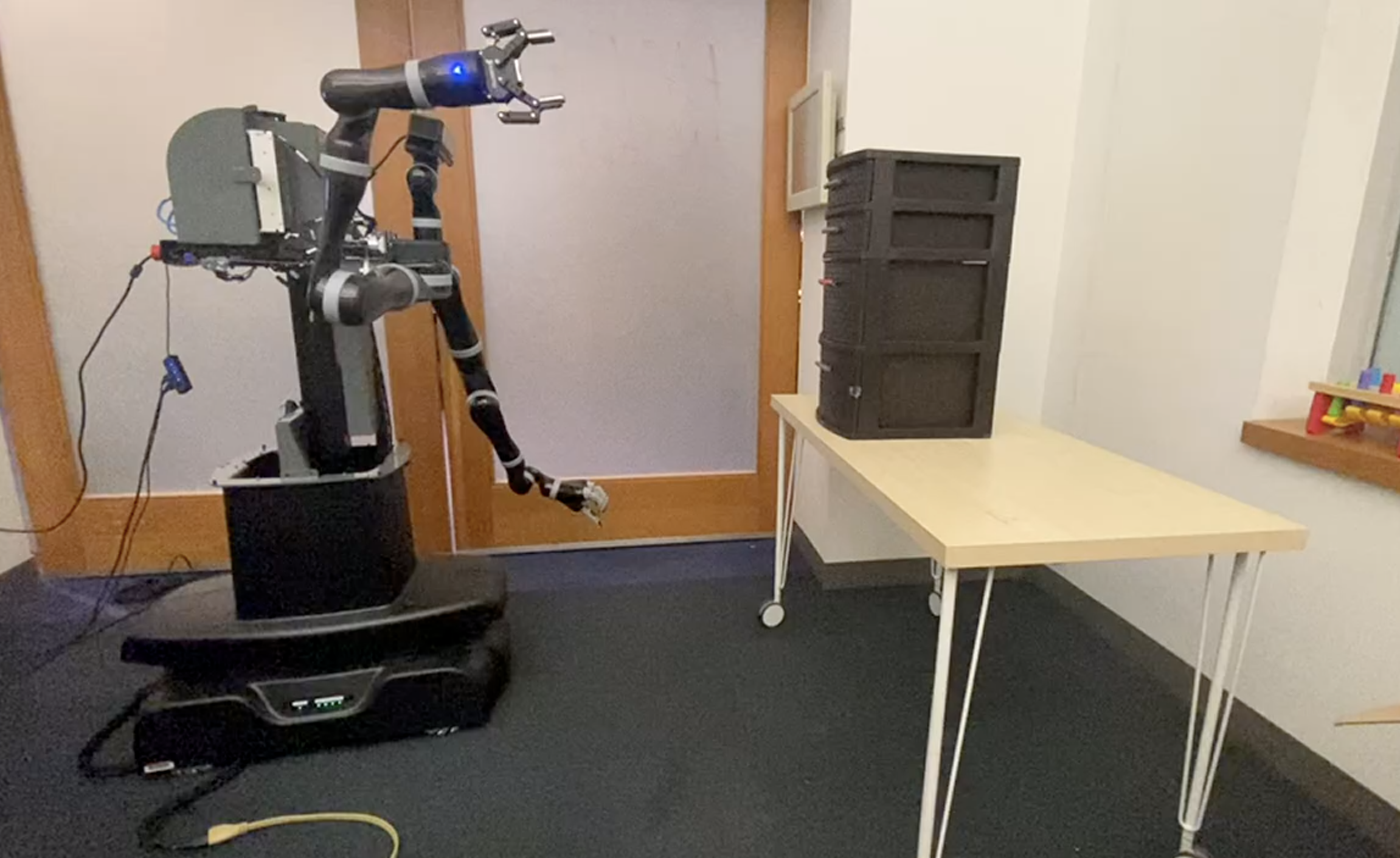}
    \end{subfigure} 
    \begin{subfigure}[b]{0.44\textwidth}
        \includegraphics[width=\textwidth]{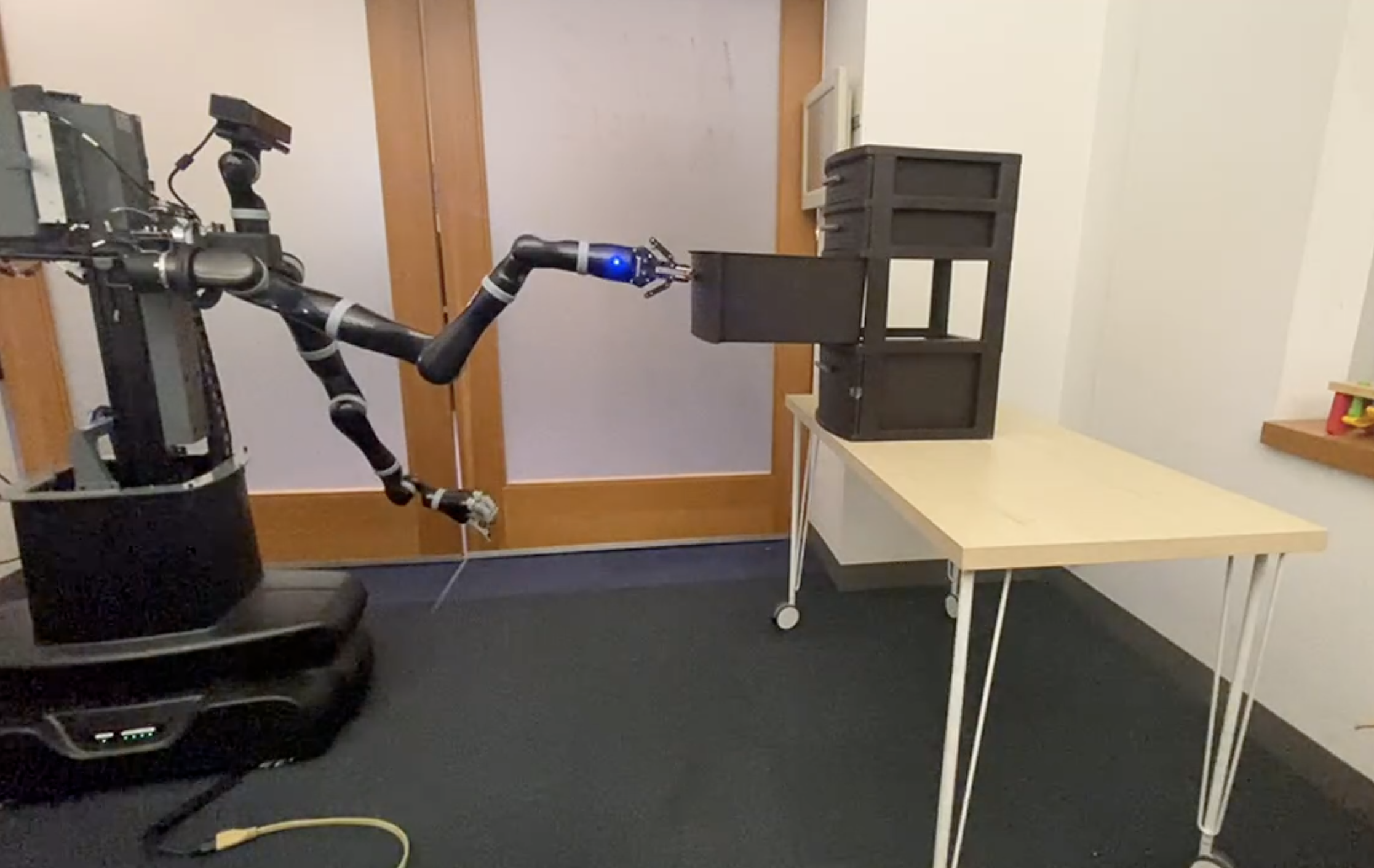}
    \end{subfigure}
     \caption{Kinova Movo opening a drawer using the drawer opening policy from ManiSkill}
     \label{fig:movo}
\end{figure} 
\section{Method}

\subsection{Environment setup}
The ManiSkill challenge is a simulated environment in the SAPIEN physics engine. The challenge uses a fictional mobile-base robot composed of a Sciurus robot body and up to two sideways-mounted panda arms. The \textit{Push chair} and \textit{Move bucket} task require a dual-arm setup, while the two cabinet tasks require only a single arm. In total, the single-arm setup has 13 degrees of freedom while the dual-arm setup has 22. The observation space includes robot joint positions and an RGB point cloud merged from three cameras that resembles an egocentric view of the scene. Additionally, task-relevant segmentation masks are provided that remove ambiguity about, for example, which cabinet to open. The action space includes joint velocities, which are clipped and scaled \footnote{let \texttt{t} be the joint limits of joint i, the evaluator script process our actions with \texttt{actions[i] = (t.upper-t.lower)/2 * actions[i] + (t.upper+t.lower)/2}}. Rewards are sparse in the environment and are only non-zero after successful completion of a task. The requirements for successful completion vary depending on the environment, but they all share a time limit of 200 simulation steps with a control frequency of 100 (~2 seconds in physical simulator time).

\subsection{Our manipulation framework}

\begin{figure*}
    \includegraphics[width=1\textwidth]{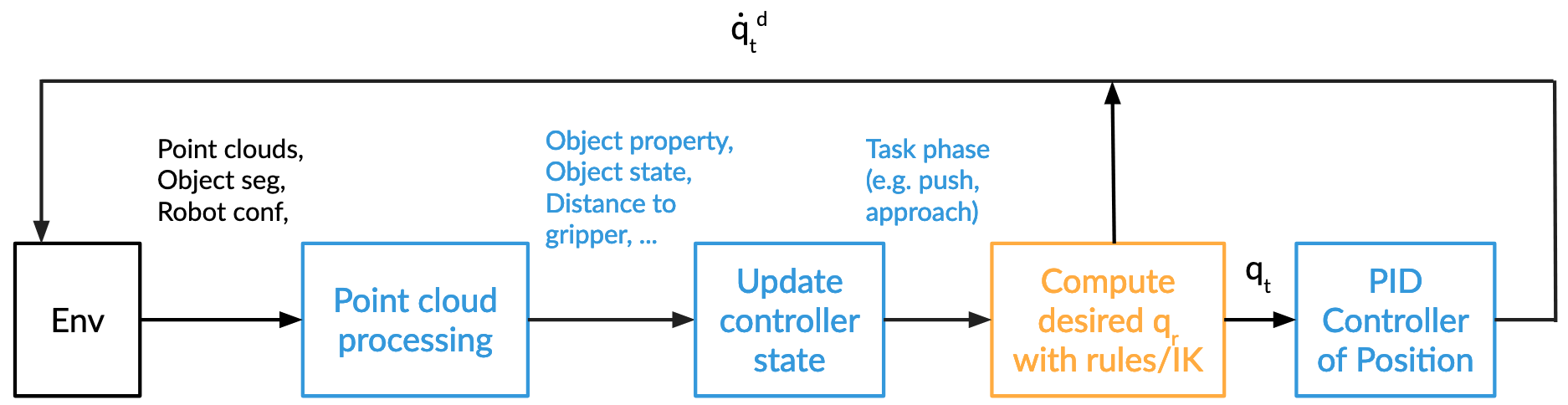}
  \caption{Manipulation pipeline. The gym simulated environment takes in actions in the form of $\dot{q}$ velocities and outputs point clouds and joint states. Our solution is a finite state machine that first processes the incoming point cloud to extract an important task-specific set of features (e.g. the bucket rim), then uses these features to compute a set of object specific affordances and the next state, then computes joint positions $q$ for these affordances, and finally calculates joint velocities $\dot{q}$ to reach the desired $q$.}
  \label{fig:diagram}
  \centering
\end{figure*}

We take a robotics approach to this problem with dedicated perception and control modules as shown in Figure~\ref{fig:diagram}. At its core, our method relies on a behavior tree that is specific to each task. This behavior tree is checked at each timestep for successful completion of an action and modifies the state if it determines an action has failed or completed successfully. The first step in our pipeline processes point clouds to obtain a set of task-relevant features. The resulting features and point cloud are passed into a state detector that detects if the particular action associated with the current behavior tree state has succeeded, failed, or is still executing. Our pipeline then performs a lookup for the desired joint positions  in the current behavior tree state. This step uses the processed point cloud features and an inverse kinematics solver to compute the desired joint positions $q^d$ from the desired end-effector positions. Lastly, our pipeline uses a $PID$ controller to calculate the desired velocity $\dot{q}^d$ from the desired joint positions $q^d$. Below we outline the specifics of each step and how they are implemented for each of the tasks.

\begin{figure*}[hbt!]
    \centering
    \begin{subfigure}[b]{0.24\textwidth}
        \includegraphics[width=\textwidth]{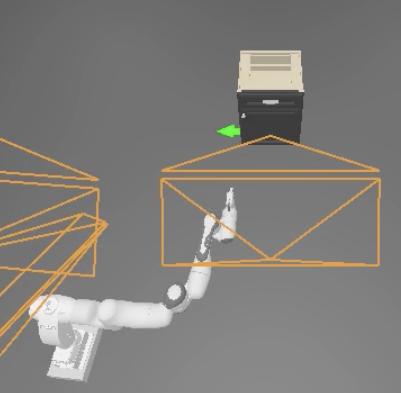}
        \caption{Initial pose}
    \end{subfigure} \hfill 
    \begin{subfigure}[b]{0.24\textwidth}
        \includegraphics[width=\textwidth]{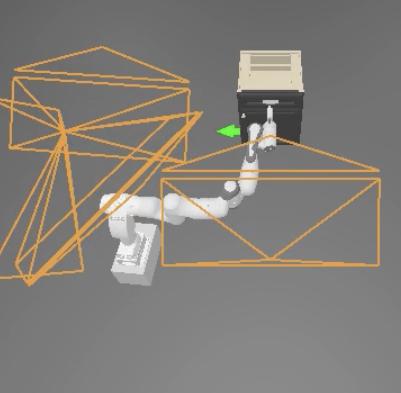}
        \caption{Pre-grasp}
    \end{subfigure} \hfill 
    \begin{subfigure}[b]{0.24\textwidth}
        \includegraphics[width=\textwidth]{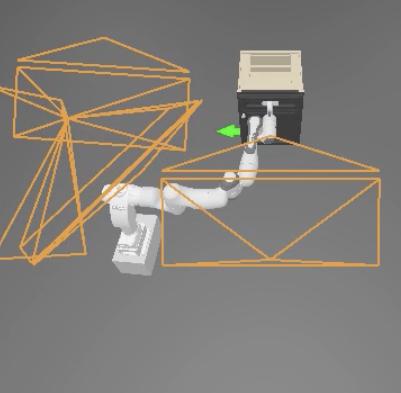}
        \caption{Grasp}
    \end{subfigure} \hfill 
    \begin{subfigure}[b]{0.24\textwidth}
        \includegraphics[width=\textwidth]{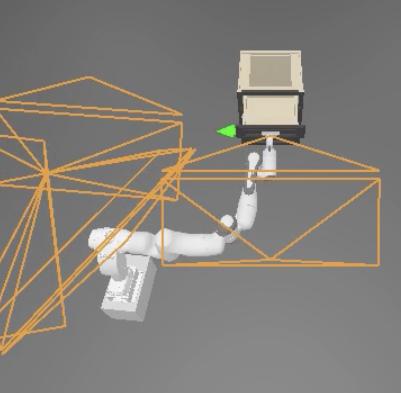}
        \caption{Open}
    \end{subfigure} 
    \caption{The policy for opening the drawer}
    \label{fig:drawer_traj}
\end{figure*}

\begin{figure*}[hbt!]
    \centering
    \begin{subfigure}[b]{0.13\textwidth}
        \includegraphics[width=\textwidth]{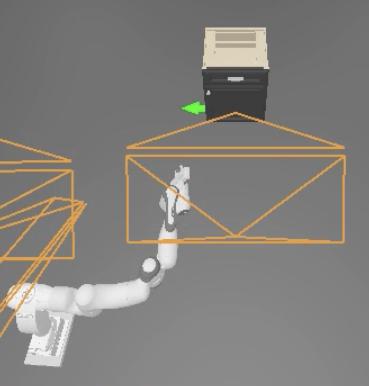}
        \caption{Initial pose}
    \end{subfigure} \hfill 
    \begin{subfigure}[b]{0.13\textwidth}
        \includegraphics[width=\textwidth]{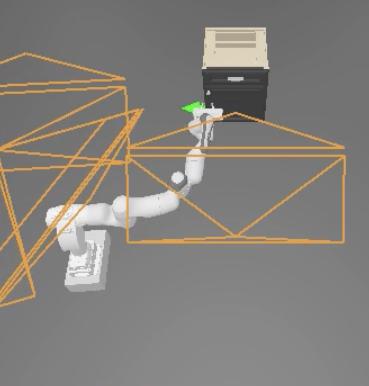}
        \caption{Pre-grasp}
    \end{subfigure} \hfill 
    \begin{subfigure}[b]{0.13\textwidth}
        \includegraphics[width=\textwidth]{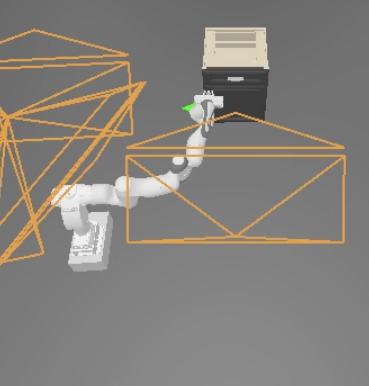}
        \caption{Grasp}
    \end{subfigure} \hfill 
    \begin{subfigure}[b]{0.13\textwidth}
        \includegraphics[width=\textwidth]{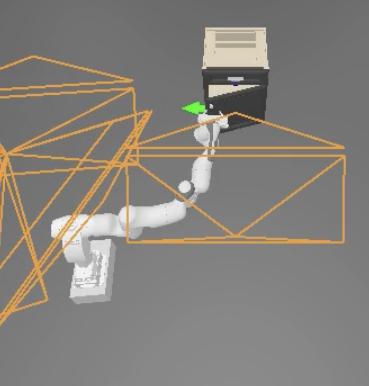}
        \caption{Open}
    \end{subfigure} \hfill 
    \begin{subfigure}[b]{0.13\textwidth}
        \includegraphics[width=\textwidth]{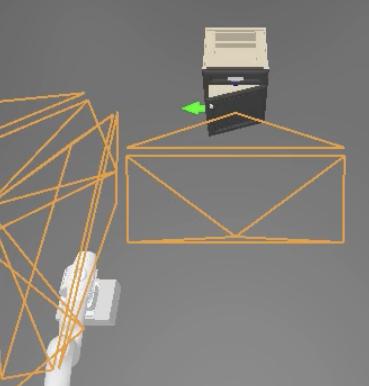}
        \caption{Rotate}
    \end{subfigure} \hfill 
    \begin{subfigure}[b]{0.13\textwidth}
        \includegraphics[width=\textwidth]{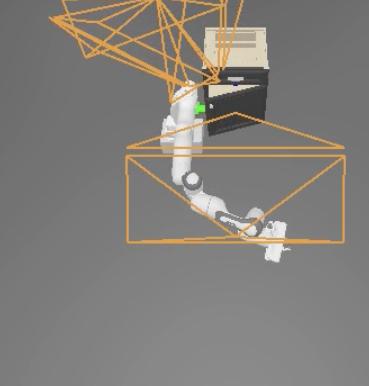}
        \caption{Pre-push}
    \end{subfigure} \hfill 
    \begin{subfigure}[b]{0.13\textwidth}
        \includegraphics[width=\textwidth]{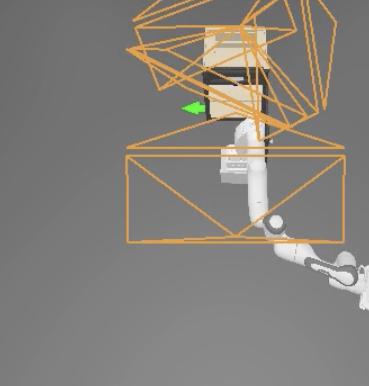}
        \caption{Push}
    \end{subfigure} \hfill 
    \caption{The policy for opening the door with a base push}
    \label{fig:door_traj}
\end{figure*}

\begin{figure*}[hbt!]
    \centering
    \begin{subfigure}[b]{0.18\textwidth}
        \includegraphics[width=\textwidth]{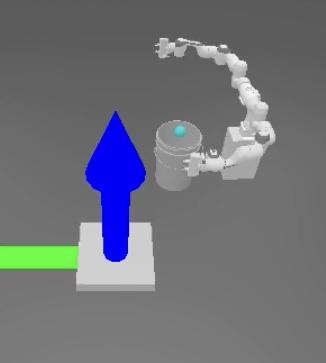}
        \caption{Initial pose}
    \end{subfigure} \hfill 
    \begin{subfigure}[b]{0.18\textwidth}
        \includegraphics[width=\textwidth]{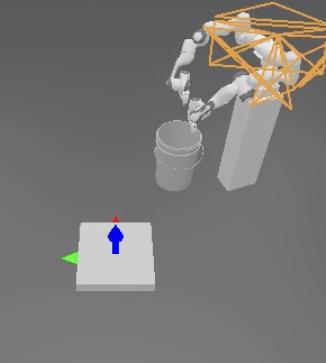}
        \caption{Pre-grasp}
    \end{subfigure} \hfill 
    \begin{subfigure}[b]{0.18\textwidth}
        \includegraphics[width=\textwidth]{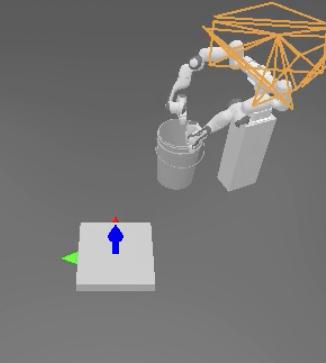}
        \caption{Grasp}
    \end{subfigure} \hfill 
    \begin{subfigure}[b]{0.18\textwidth}
        \includegraphics[width=\textwidth]{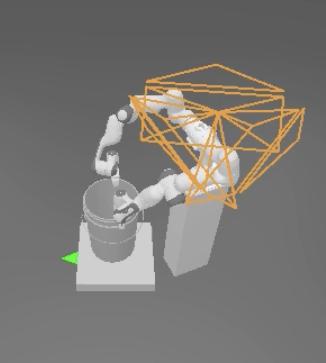}
        \caption{Move to drop}
    \end{subfigure} \hfill 
    \begin{subfigure}[b]{0.18\textwidth}
        \includegraphics[width=\textwidth]{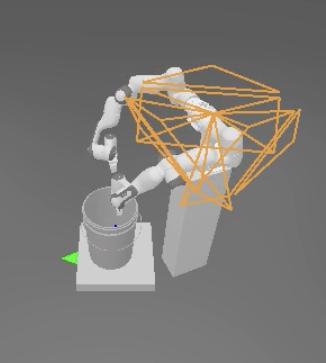}
        \caption{Drop}
    \end{subfigure} 
    \caption{The policy for grasping the bucket with two arms}
    \label{fig:bucket_prehensile_traj}
\end{figure*}

\begin{figure*}[hbt!]
    \centering
    \begin{subfigure}[b]{0.18\textwidth}
        \includegraphics[width=\textwidth]{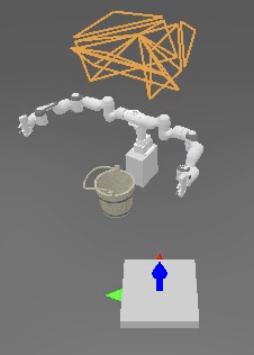}
        \caption{Initial pose}
    \end{subfigure} \hfill 
    \begin{subfigure}[b]{0.18\textwidth}
        \includegraphics[width=\textwidth]{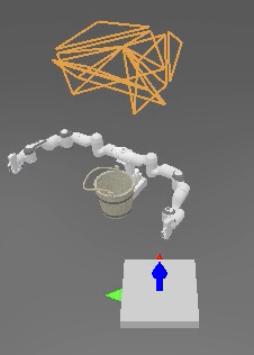}
        \caption{Pre-hug}
    \end{subfigure} \hfill 
    \begin{subfigure}[b]{0.18\textwidth}
        \includegraphics[width=\textwidth]{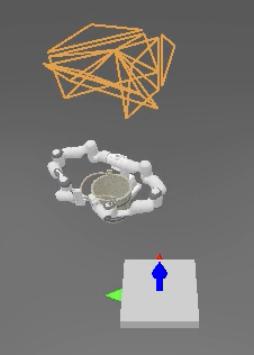}
        \caption{Hug}
    \end{subfigure} \hfill 
    \begin{subfigure}[b]{0.18\textwidth}
        \includegraphics[width=\textwidth]{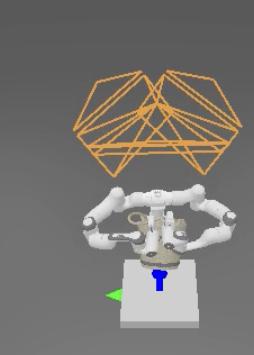}
        \caption{Drop pose}
    \end{subfigure} \hfill 
    \begin{subfigure}[b]{0.18\textwidth}
        \includegraphics[width=\textwidth]{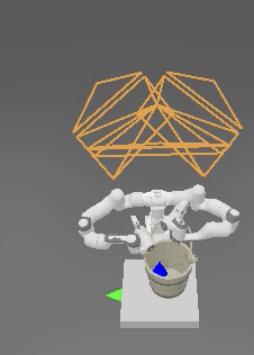}
        \caption{Drop}
    \end{subfigure} 
    \caption{The policy for hugging the bucket with two arms}
    \label{fig:bucket_hug_traj}
\end{figure*}

\begin{figure*}[hbt!]
    \centering
    \begin{subfigure}[b]{0.24\textwidth}
        \includegraphics[width=\textwidth]{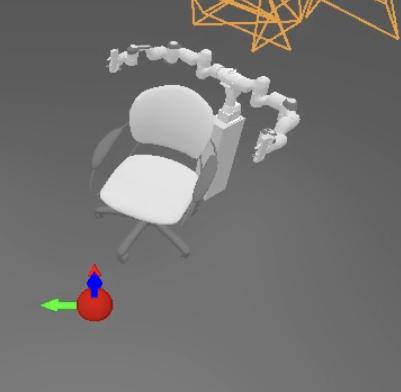}
        \caption{Initial pose}
    \end{subfigure} \hfill 
    \begin{subfigure}[b]{0.24\textwidth}
        \includegraphics[width=\textwidth]{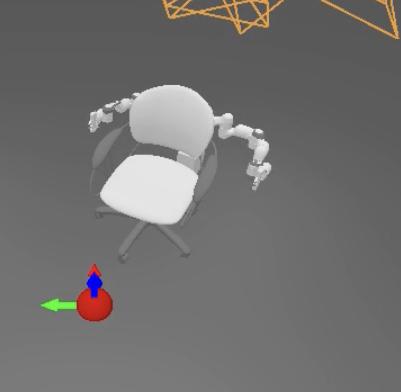}
        \caption{Move to pre-push}
    \end{subfigure} \hfill 
    \begin{subfigure}[b]{0.24\textwidth}
        \includegraphics[width=\textwidth]{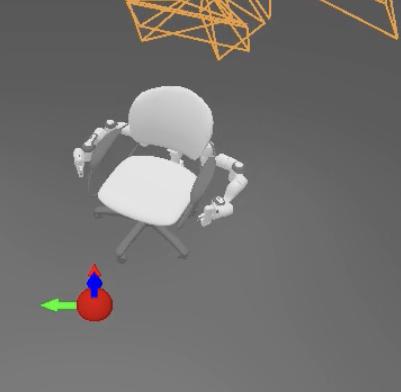}
        \caption{Push close to target}
    \end{subfigure} \hfill 
    \begin{subfigure}[b]{0.24\textwidth}
        \includegraphics[width=\textwidth]{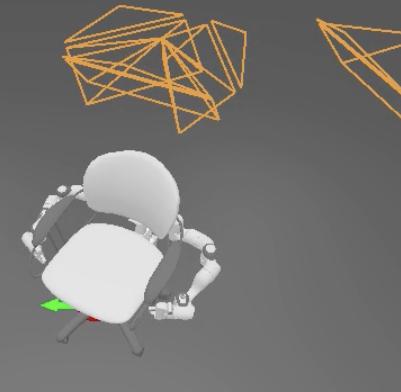}
        \caption{Nudge towards target}
    \end{subfigure} 
    \caption{The policy for hug-pushing and nudging the chair with two arms}
    \label{fig:chair_traj}
\end{figure*}

\subsection{point cloud processing}
\begin{figure}[hbt]
    \centering
    \begin{subfigure}[b]{0.23\textwidth}
        \includegraphics[width=0.5\textwidth]{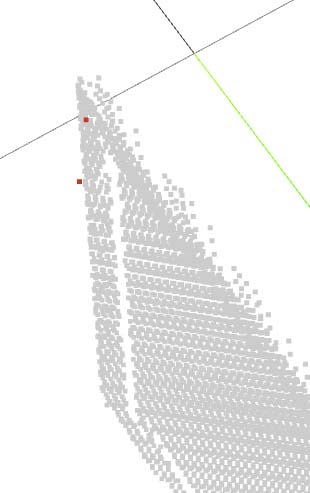}
        \subcaption{Sparse points in the first frame}
    \end{subfigure} \hfill 
    \begin{subfigure}[b]{0.23\textwidth}
        \includegraphics[width=0.5\textwidth]{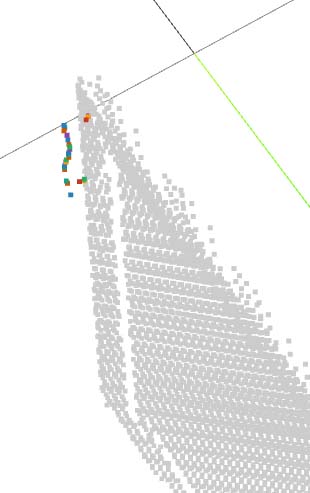}
        \subcaption{Accumulated points in the first 10 frames}
    \end{subfigure} \hfill 
    \begin{subfigure}[b]{0.23\textwidth}
        \includegraphics[width=0.5\textwidth]{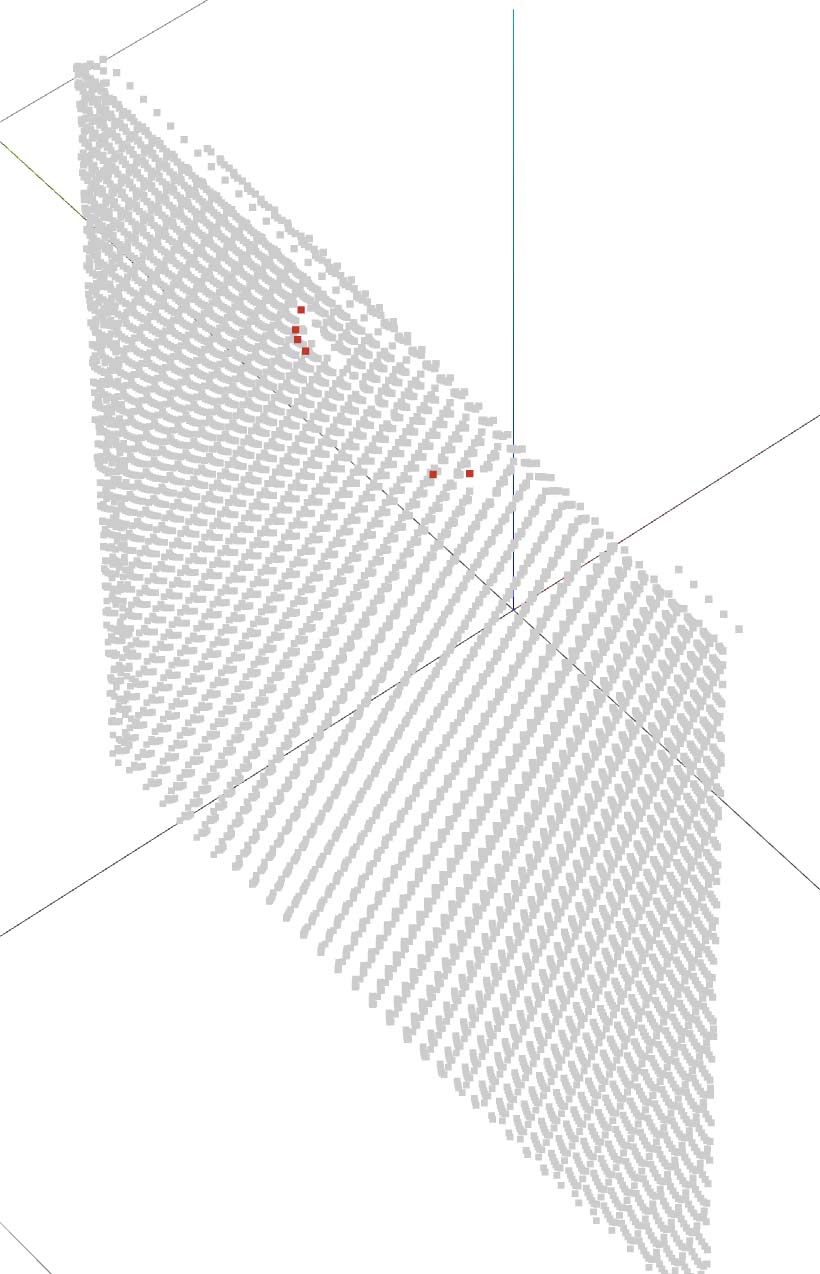}
        \subcaption{Sparse points in the first frame}
    \end{subfigure} \hfill 
    \begin{subfigure}[b]{0.23\textwidth}
        \includegraphics[width=0.5\textwidth]{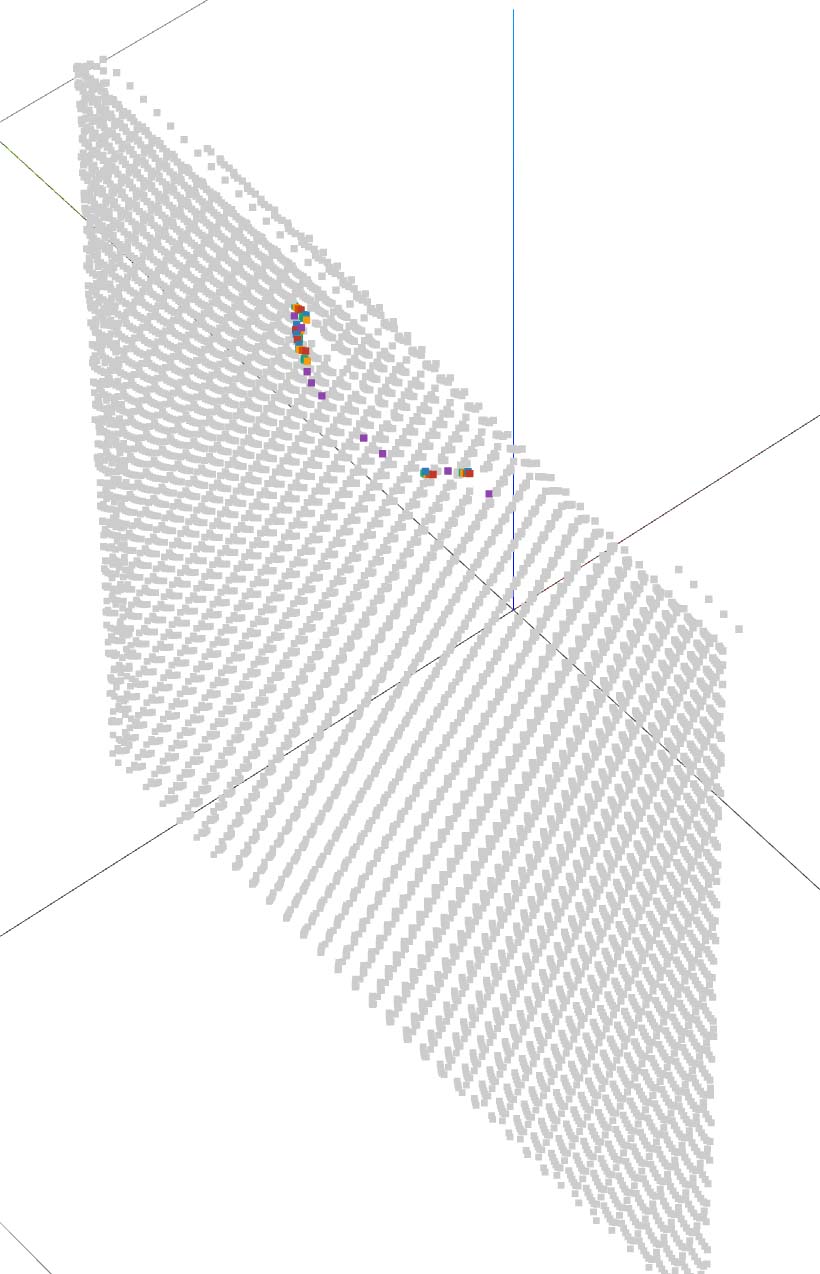}
        \subcaption{Accumulated points in the first 10 frames}
    \end{subfigure} 
    
    \caption{Comparing a single point cloud and seven point clouds representation of the door handles. The rainbow color scheme encodes the sequence with which the points were registered.}
    \label{fig:merge}
\end{figure}

\noindent
\textbf{Calculating grasps on a handle}:
To calculate grasps on a handle, we use an antipodal grasp estimator. We initially wanted to filter out bad grasps using a learned grasp scoring network, but we found that horizontal or vertical grasps near the center of the handle worked well enough, and our failure cases indicated that the performance bottleneck lies elsewhere. Instead, we simply filtered by horizontal and vertical grasps and sorted by proximity to the handle centroid. We also found it helpful to merge door handle points from multiple time steps for a better estimate, as some of the initial observations are too sparse (as shown in Figure~\ref{fig:merge}). For door opening, we made sure that we have at least 50 points or have observed ten steps, whichever comes first, before we compute the centroid. We used heuristics to decide the orientation of the grasp (horizontal or vertical) based on the shape of the handle. 
 
 \noindent
 \textbf{Detecting open cabinet }:
 Our behavior tree for cabinet tasks relies on an evaluation of whether the door/drawer is open or closed. We do this by calculating the difference in x-position from the initial x-position of the segmented door/drawer region.

\noindent
\textbf{Estimating the bucket rim}:
To estimate the bucket rim, we used a modified RANSAC. We can't simply take the highest points as the bucket rim because the handle sometimes goes over the rim. To avoid this, we find the z-slices that contain at least a threshold of points, and then select the z-slice with the highest z-value. The rim is then all the points within that z-plane.

\begin{figure}[hbt]
    \centering
    \begin{subfigure}[b]{0.44\textwidth}
        \includegraphics[width=\textwidth]{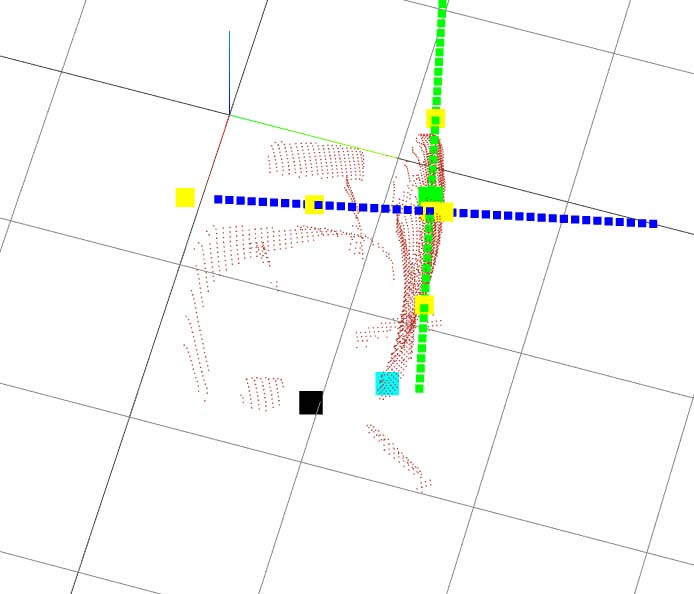}
    \end{subfigure} \hfill 
    \begin{subfigure}[b]{0.44\textwidth}
        \includegraphics[width=\textwidth]{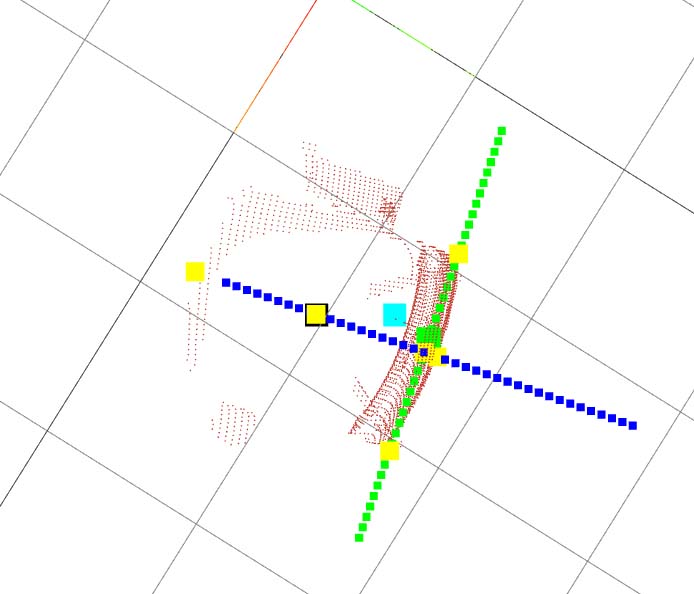}
    \end{subfigure} 
    
    \caption{PCA analysis on the chair point cloud. The green line shows the main axis of the back of the chair, while the blue axis shows the secondary axis. Points are projected onto the lines to find the bounding points, marked in yellow. The estimated center is marked in black, which is more accurate (MSE=0.038) than simply taking the mean of all points in the whole chair (MSE=0.11), as marked in cyan.}
    \label{fig:pca}

\end{figure}

\noindent
\textbf{Estimating the chair center}: 
Parts of the chair are occluded by itself because the cameras are mounted on the robot, which stands behind the chair when initialized. Thus it would be inaccurate to estimate the center of one or multiple point clouds. Instead, we rely on the assumption that the upper body of each chair (including and above the cushion) has symmetry. We first estimate the orientation of the back of the chair on the z-plane using Principle Component Analysis (PCA), then we identify the bounding box of the chair based on that axis. We calculate the center of a chair as the center of that bounding box (as shown in Figure ~\ref{fig:pca}).

\subsection{Inverse Kinematics} 
We experimented with two optimization-based algorithms for inverse kinematics: the built-in PyBullet and Drake IK tools. We decided to use Drake in the final version because PyBullet doesn't respect joint limits and cannot handle two desired gripper link poses (Necessary for the prehensile branch of the bucket move task). Additionally, Drake allowed us to add additional terms to the optimization that proved invaluable. Namely, we added a \textsc{AddPointDistanceConstraint} that forced the base of the robot to be a threshold distance from the gripper poses. This was necessary because otherwise, the arm would typically collide with the base, and the only alternative was motion planning. 

\subsection{Task-specific behavior trees}

\begin{figure}[hbt!]
    \centering
    \includegraphics[width=0.9\textwidth]{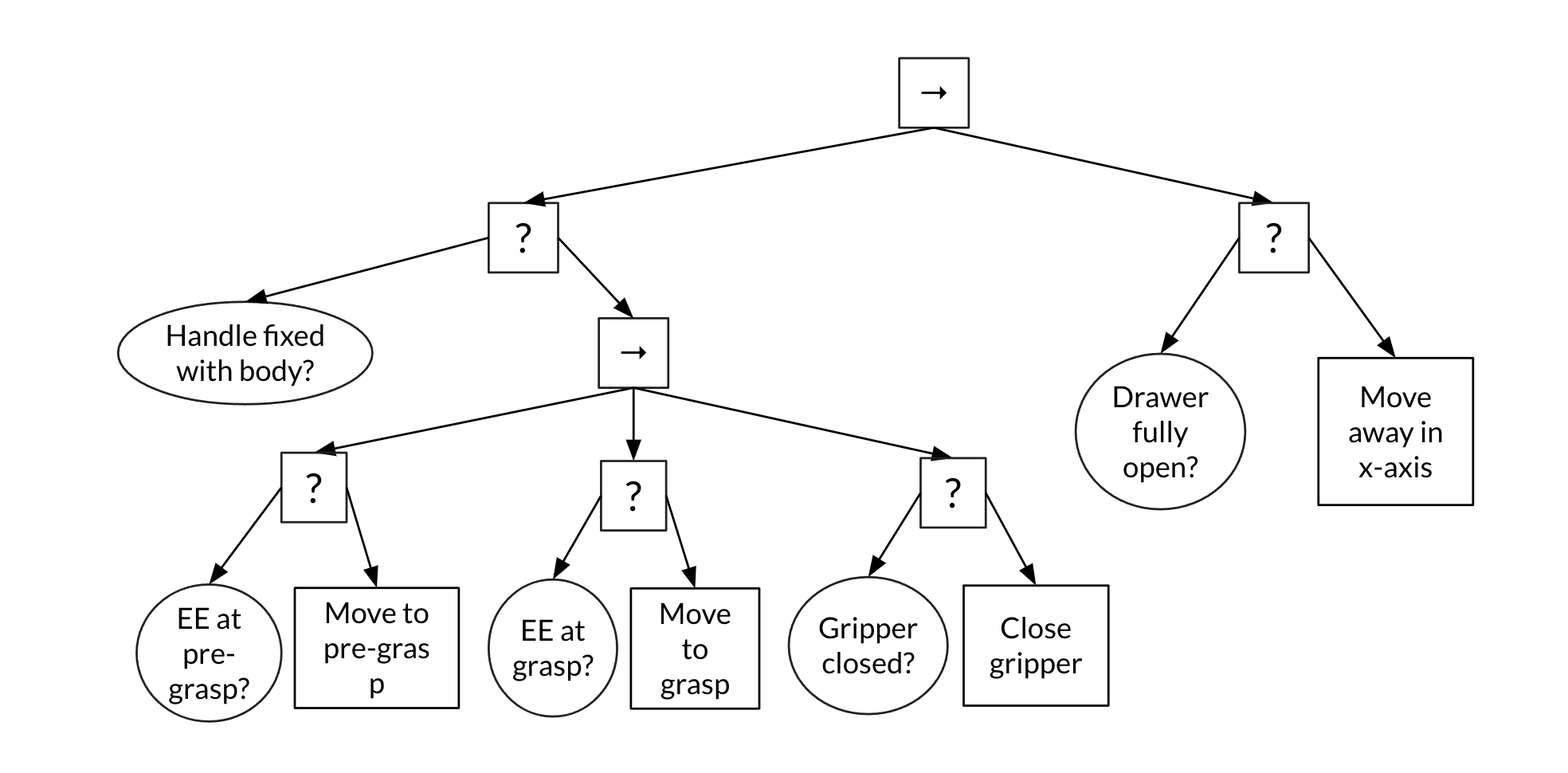}
    \caption{Behavior Decision Diagram for the \textit{Open Cabinet Drawer} task}
    \label{fig:bdd1}
\end{figure}

\noindent
\textbf{Open Cabinet Drawer}:
The behavior tree for the \textit{Cabinet Drawer} task (shown in Figure ~\ref{fig:bdd1}) moves to a pre-grasp pose, moves to a grasp pose, grasps, and then moves the base in the -z direction while checking the open status of the drawer. An example trajectory can be seen in Figure~\ref{fig:drawer_traj}.

\begin{figure}[hbt!]
    \centering
    \includegraphics[width=0.9\textwidth]{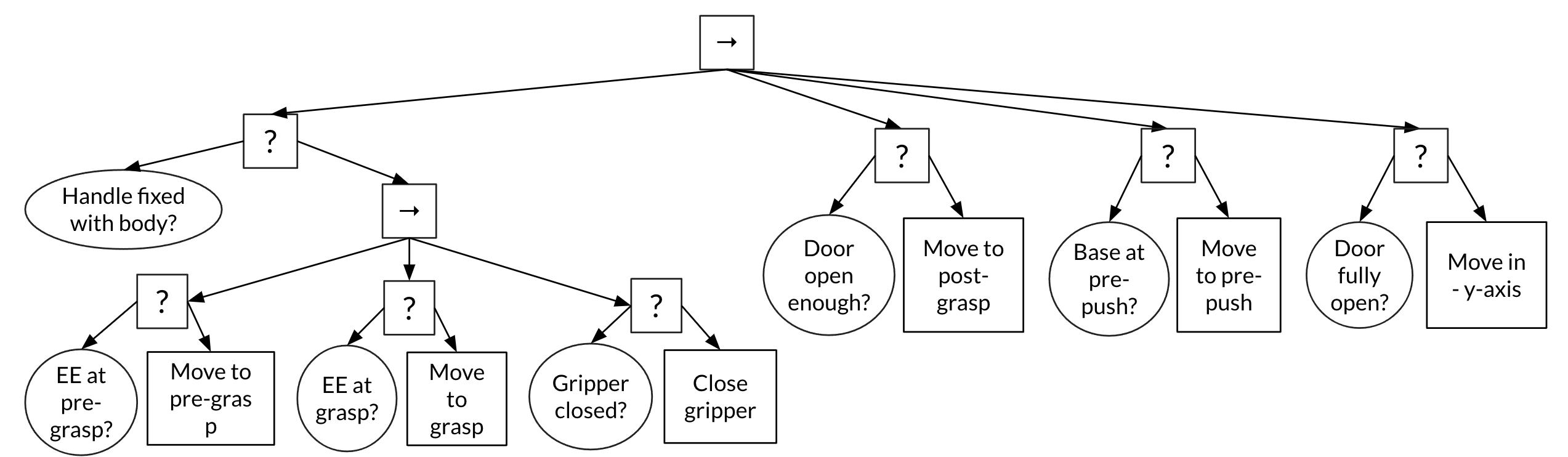}
    \caption{Behavior Decision Diagram for the \textit{Open Cabinet Door} task}
    \label{fig:bdd2}
\end{figure}

\begin{figure*}[hbt]
    \centering
    \begin{subfigure}[b]{0.32\textwidth}
        \includegraphics[width=\textwidth]{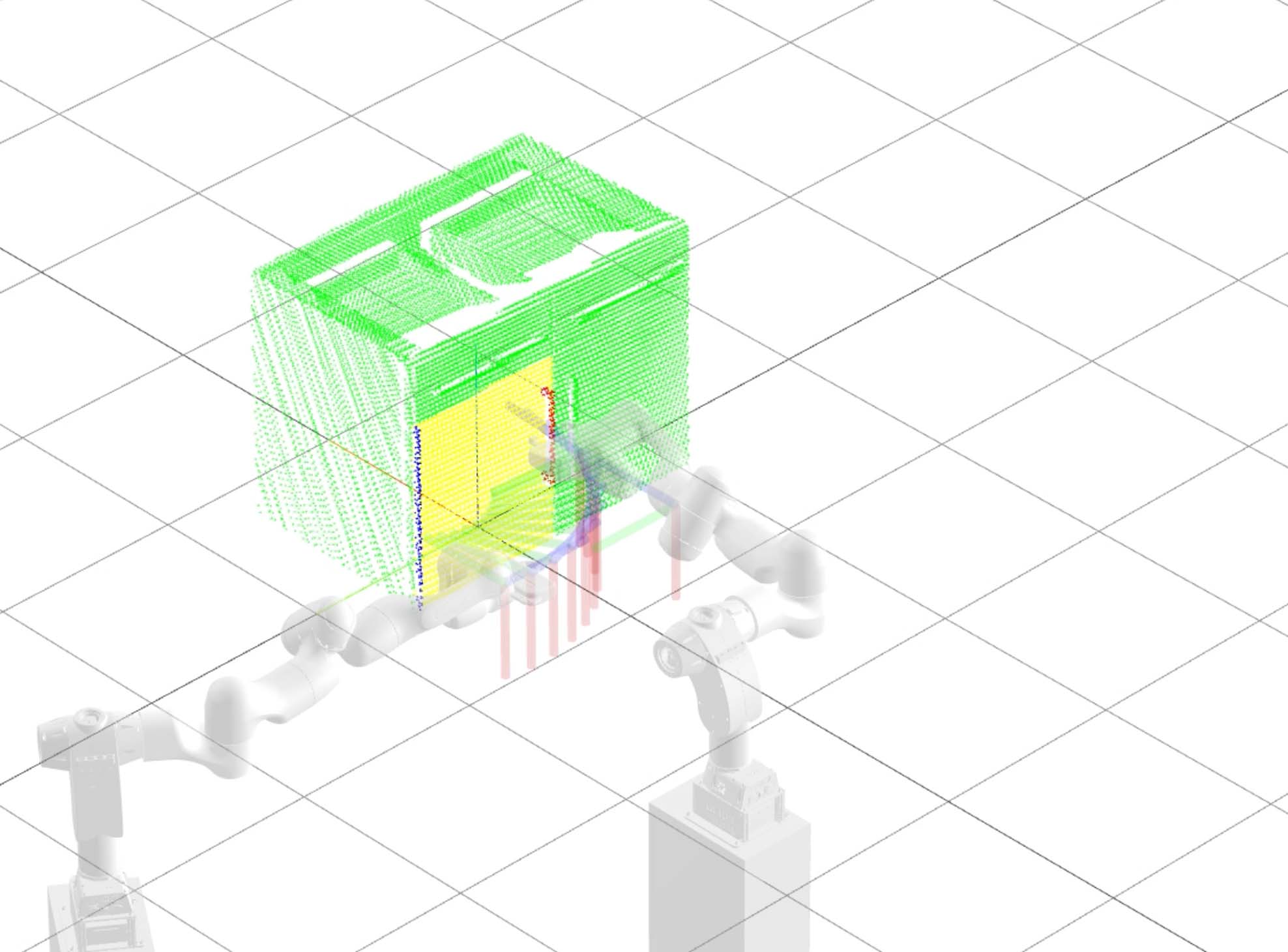}
    \end{subfigure} \hfill 
    \begin{subfigure}[b]{0.32\textwidth}
        \includegraphics[width=\textwidth]{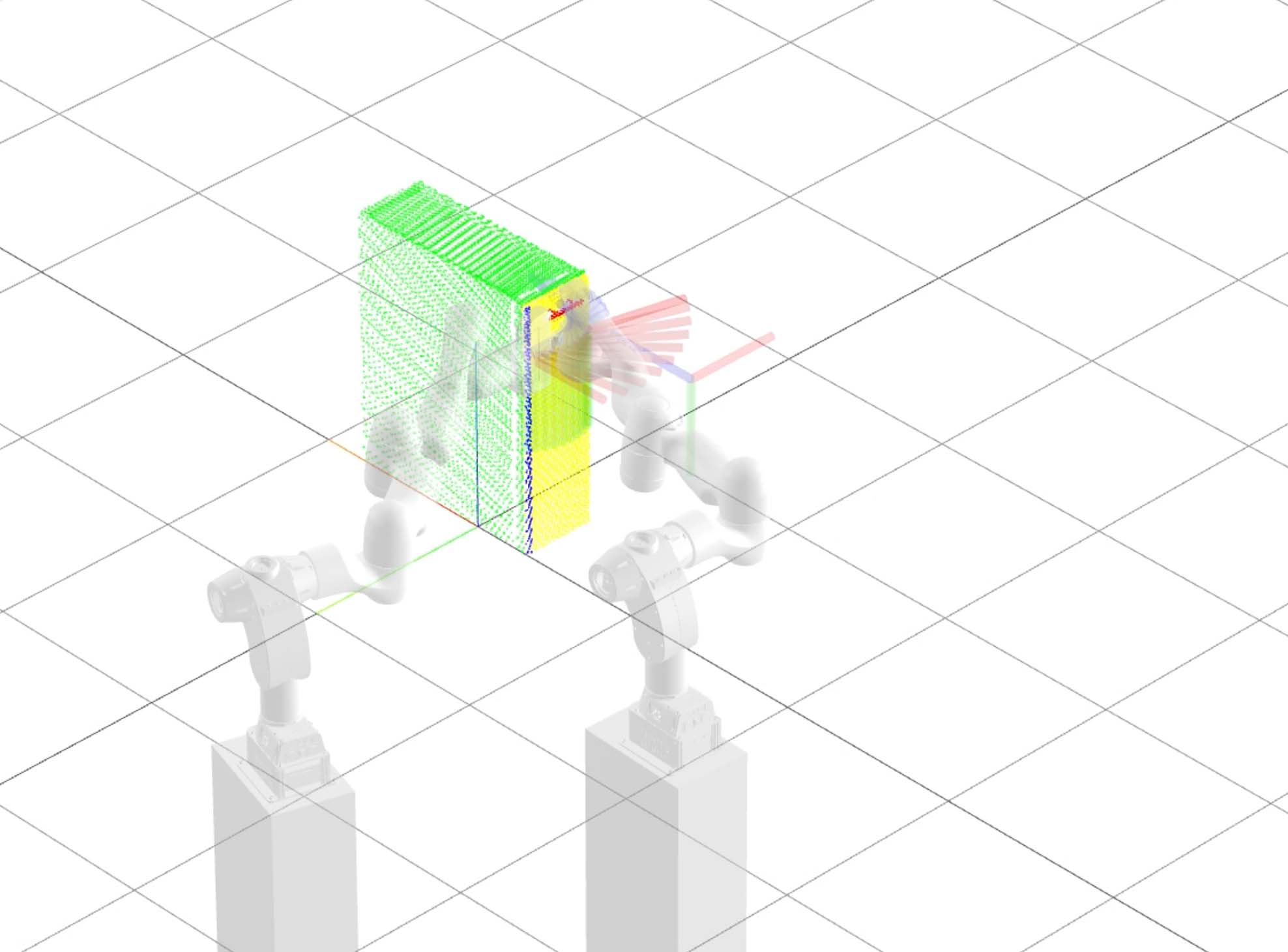}
    \end{subfigure} \hfill 
    \begin{subfigure}[b]{0.32\textwidth}
        \includegraphics[width=\textwidth]{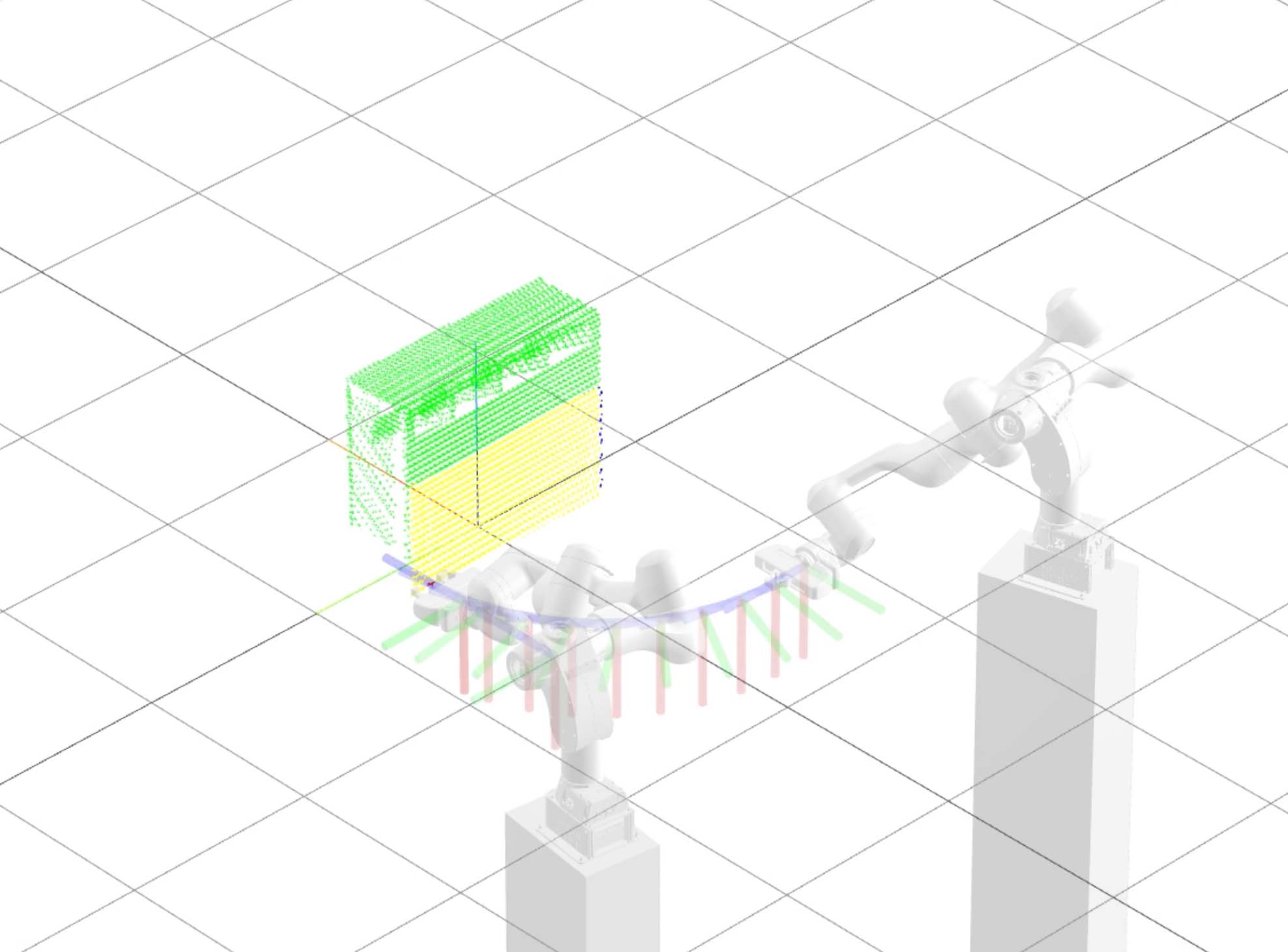}
    \end{subfigure}
    \caption{Trajectories for opening a door consists of a pre-grasp pose and ten key poses in the circular motion}
    \label{fig:door_traj_circ}
\end{figure*}

\noindent
\textbf{Open Cabinet Door}:
The behavior tree for the \textit{Cabinet Door} task (shown in Figure ~\ref{fig:bdd2}) moves to a pre-grasp pose, moves to a grasp pose, grasps, and then moves the base diagonally in the -z, -x direction for right opening doors and -z, x direction for left opening doors while checking the open status of the door. After the door is partially open, it moves the base to pry the door open further. An example trajectory can be seen in Figure~\ref{fig:door_traj}. We also experimented with using a circular gripper path to open the door completely. Although this method works well, the time constraints did not allow us to add enough intermediate gripper poses for a smooth gripper path (Figure~\ref{fig:door_traj_circ}).

\begin{figure}[hbt!]
    \centering
    \includegraphics[width=0.9\textwidth]{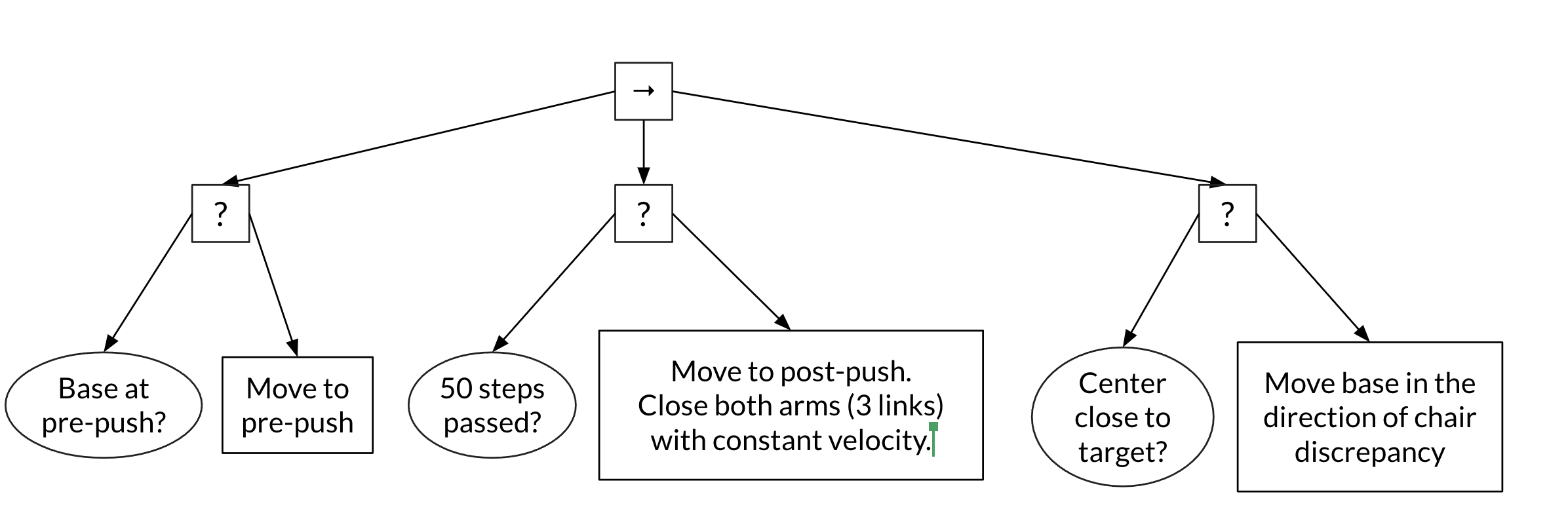}
    \caption{Behavior Decision Diagram for the \textit{Push Chair} task}
    \label{fig:bdd3}
\end{figure}

\noindent
\textbf{Move Bucket}:
The behavior tree for the \text{Move Bucket} task (shown in Figure ~\ref{fig:bdd3}) first tests the thickness of the bucket rim. If it is too thick to grasp, it follows the ``hugging'' branch. The hugging branch simply moves the base to in front of the bucket, ``hug'' by closing both arms around the bucket, lifting the bucket, moving to the target platform, and opening its arms ``unhugs'' the bucket. An example trajectory from the hugging branch is in Figure~\ref{fig:bucket_hug_traj}. Otherwise, if the bucket rim is sufficiently thin, the behavior tree follows the prehensile branch. This branch moves the base to in front of the bucket, grasps the bucket with both grippers, lifts the bucket by increasing the base height, moves to the target region, and drops the bucket. An example trajectory from the prehensile branch is in Figure~\ref{fig:bucket_prehensile_traj}.

\begin{figure}[hbt!]
    \centering
    \includegraphics[width=0.9\textwidth]{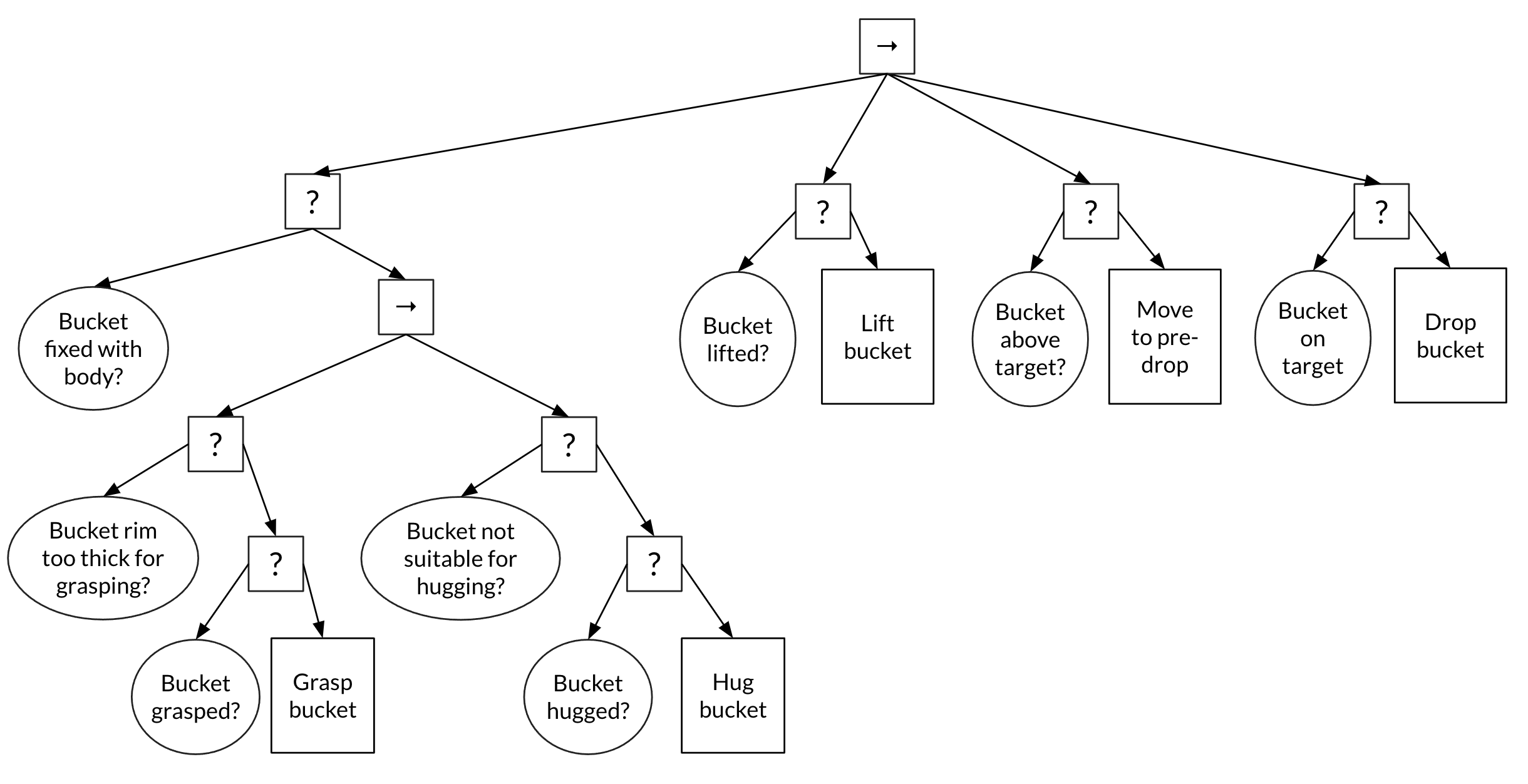}
    \caption{Behavior Decision Diagram for the \textit{Move Bucket} task}
    \label{fig:bdd4}
\end{figure}

\noindent
\textbf{Push Chair}:
The behavior tree for the \text{Push Chair} task (shown in Figure~\ref{fig:bdd4}) moves the base to be in front of the chair, ``hugs'' the chair using the same hugging policy from bucket picking, and then moves the base such that the hair center is at the world center. It then enters a ``nudging'' phase in which it continually estimates the chair center and slightly adjusts its position accordingly. An example trajectory can be found in Figure~\ref{fig:chair_traj}.

\noindent
\subsection{PID Controller}
We used a custom PID controller using the equations from class. We additionally added a low-pass filter ($f_s=100$Hz, $f_c=40$Hz) to the output to avoid large jumps in output velocity. We found that the PID parameters needed to be different for different components of the robot due to the different scales of position differences and the strict time limit. For base translation and height control we used $k_p=20$, $k_i=0.5$, $k_d=0.0$. For arm position control we used $k_p=10$, $k_i=2.0$, $k_d=0.0$. Lastly for base rotation control we used $k_p=0.5$, $k_i=0.2$, $k_d=0.0$. For some reason, we believe because of the low control frequency $k_d>0.0$ led to significant instability in the form of jittering.

\section{Results}

\subsection{Evaluation of task success}

\begin{table*}[ht!]
\centering
\begin{tabular}{lccccccccccc}\toprule
Approaches & \multicolumn{2}{c}{BC} & \multicolumn{2}{c}{BCQ (Offline RL)} & \multicolumn{2}{c}{TD3+BC} & \multicolumn{2}{c}{Ours}
\\\cmidrule(lr){2-3}\cmidrule(lr){4-5}\cmidrule(lr){6-7}\cmidrule(lr){8-9}
Tasks $\backslash$ Split               & Train & Test & Train & Test  & Train & Test  & Train & Test  \\\midrule
OpenCabinetDrawer   & $0.37$ & $0.12$ & $0.22$ & $0.11$ & $0.18$ & $0.10$ & 0.972 & 0.988 \\
OpenCabinetDoor     & $0.30$ & $0.11$ & $0.16$ & $0.04$ & $0.13$ & $0.04$ & 0.472 & 0.28 \\
PushChair           & $0.18$ & $0.08$ & $0.11$ & $0.12$ & $0.12$ & $0.08$ & 0.732 & 0.728 \\
MoveBucket          & $0.15$ & $0.08$ & $0.08$ & $0.12$ & $0.05$ & $0.03$ & 0.772 & 0.572 \\\bottomrule
\end{tabular}
\caption{Training and testing scores for each of our solutions according to official evaluation, compared with the ManiSkill benchmark.}
\label{tab:1}
\end{table*}

We compare our task success with an Behavior Cloning (BC) baseline provided by the ManiSkill challenge. Their architecture uses PointNet and Transformer networks, trained on 100 trajectories for each training environment with 150k gradient steps\footnote{They provided four baselines, we took the ones with the best performance from \url{https://github.com/haosulab/ManiSkill-Learn}}. We retrieved the evaluation results of our solutions from the ManiSkill official website. See \ref{tab:1} for results.

\subsection{Demonstration on real robot}
We applied the control framework to get a dual-arm Kinova Movo robot to open a cabinet drawer. We used red color tape to coat the designated handle and used Drake for IK (Figure~\ref{fig:movo} ). See a video demo at \url{https://streamable.com/rntmv2}

\section{Discussion}
Our results show that our approach is well-suited for designing generalizable manipulation skills. While it's not the only way to construct manipulation skills, we feel this approach lends itself to high generalizability and interpretability. We can look at the failure cases and adjust our policy to address those failures. This challenge was primarily built for machine learning solutions, evidenced by the low-frequency joint observations, limited time-horizon, and narrow initial state distributions. Despite this, we were pleased to find that standard techniques in robotic manipulation, in most cases, outperformed the machine learning solutions. Our solution is also directly transferable to the real world. It runs no risk of overfitting to the simulator and automatically generalizes to arbitrary scenes that might contain these articulated objects without domain randomization or dataset augmentation. 
The policies we developed for this challenge significantly improve with more simulator time as they can enter the policy's replanning branch on subgoal failure. Additionally, a higher frequency controller akin to those on real robotic systems would enable us to implement circular and linear gripper trajectories without moving the base and include motion planning for added robustness in real-world applications.

\bibliography{gpl_iclr2022_conference}
\bibliographystyle{gpl_iclr2022_conference}
\end{document}

%% file: math_commands.tex
\usepackage{amsmath,amsfonts,bm}

\def\eqref#1{equation~\ref{#1}}

\def\1{\bm{1}}

\DeclareMathAlphabet{\mathsfit}{\encodingdefault}{\sfdefault}{m}{sl}
\SetMathAlphabet{\mathsfit}{bold}{\encodingdefault}{\sfdefault}{bx}{n}

